\documentclass[10pt,twocolumn,letterpaper]{article}

\usepackage[pagenumbers]{iccv}

\definecolor{iccvblue}{rgb}{0.21,0.49,0.74}
\usepackage[pagebackref,breaklinks,colorlinks,allcolors=iccvblue]{hyperref}
\usepackage{enumitem}
\setlist[itemize]{topsep=0pt, itemsep=0pt, partopsep=0pt}

\title{LVC: A Lightweight Compression Framework for Enhancing VLMs in Long Video Understanding}

\author{Ziyi Wang$^{1,2}$\footnotemark[1], Haoran Wu$^{1}$\footnotemark[1], Yiming Rong$^{1,2}$\footnotemark[1], Deyang Jiang$^{1,2}$, Yixin Zhang$^{1,2}$\\
Yunlong Zhao$^{1,2}$, Shuang Xu$^{1}$, Bo Xu$^{1}$\\
$^{1}$ The Key Laboratory of Cognition and Decision Intelligence for Complex Systems,\\ Institute of Automation, Chinese Academy of Sciences\\
$^{2}$ School of Artificial Intelligence, University of Chinese Academy of Sciences\\
{
\tt\small \{wangziyi2022, wuhaoran2018, rongyiming2022, jiangdeyang2022\}@ia.ac.cn}, \\
\tt\small \{zhangyixin2024, zhaoyunlong2020, shuang.xu, xubo\}@ia.ac.cn
}

\begin{document}
\maketitle

{\renewcommand{\thefootnote}{\fnsymbol{footnote}}
    \footnotetext[1]{Corresponding author.}}

\begin{abstract}
    Long video understanding is a complex task that requires both spatial detail and temporal awareness. 
    While Vision-Language Models (VLMs) obtain frame-level understanding capabilities through multi-frame input, they suffer from information loss due to the sparse sampling strategy.
    In contrast, Video Large Language Models (Video-LLMs) capture temporal relationships within visual features but are limited by the scarcity of high-quality video-text datasets.
    To transfer long video understanding capabilities to VLMs with minimal data and computational cost, we propose Lightweight Video Compression (LVC), a novel method featuring the Query-Attention Video Compression mechanism, which effectively tackles the sparse sampling problem in VLMs.
    By training only the alignment layer with 10k short video-text pairs, LVC significantly enhances the temporal reasoning abilities of VLMs. 
    Extensive experiments show that LVC provides consistent performance improvements across various models, including the InternVL2 series and Phi-3.5-Vision. 
    Notably, the InternVL2-40B-LVC achieves scores of 68.2 and 65.9 on the long video understanding benchmarks MLVU and Video-MME, respectively, with relative improvements of 14.6\% and 7.7\%.
    The enhanced models and code will be publicly available soon.
\end{abstract}    
\section{Introduction}
\begin{figure*}[h]
  \centering
   \includegraphics[width=\linewidth]{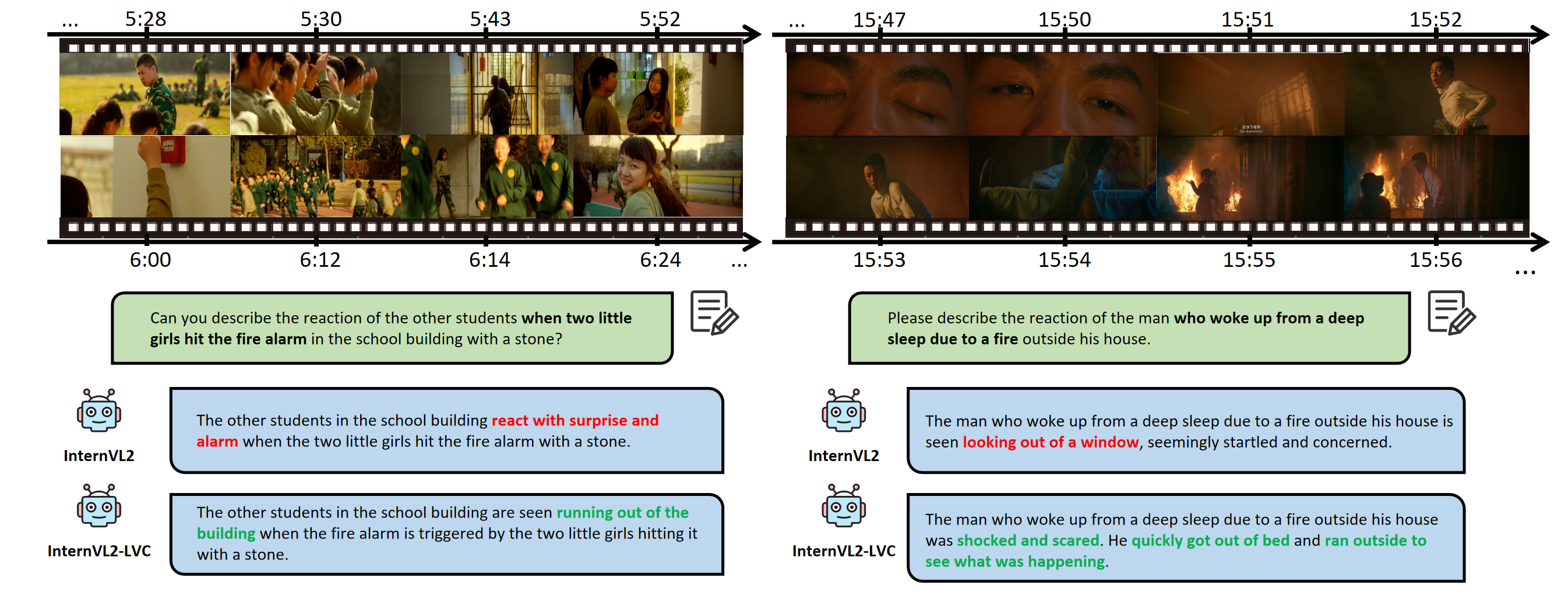} 
   \caption{Comparison examples of InternVL2-40B and InternVL2-40B-LVC. \textcolor{red}{Red} text indicates errors in InternVL2-40B caused by its sparse sampling strategy failing to capture temporal details, while \textcolor{green}{green} text highlights corrections made by InternVL2-40B-LVC.}
   \label{Figure1}
\end{figure*}

The rapid advancement of large language models (LLMs) has driven a paradigm shift in video understanding research, transitioning from conventional vision-centric approaches~\citep{tran2015learning, feichtenhofer2020x3d, sun2019videobert, bertasius2021space} to LLM-based frameworks that leverage cross-modal alignment capabilities. 
This LLM-driven revolution manifests in two predominant architectures: Video-LLMs~\citep{lin2023video, jin2024chat, ataallah2024minigpt4} pre-trained on video-text-paired data and VLMs~\citep{chen2024internvl, li2024llava, wang2024qwen2} centered on image-text alignment~\cite{li2023blip, liu2023visual}.

Video-LLMs establish video-text alignment~\cite{ren2024timechat, huang2024vtimellm}, projecting sequential frame features into LLM's text space. 
While Video-LLMs demonstrate superior temporal awareness through video pre-training, their practical implementation faces two key bottlenecks: limited availability of high-quality video-text datasets~\cite{maaz2023video, luo2023valley} and expensive computational costs.
In contrast, by processing temporally sampled frames as discrete visual inputs, VLMs demonstrate surprising competence in short video understanding tasks~\cite{caba2015activitynet, xu2016msr, li2024mvbench}.
Their strong spatial reasoning abilities, derived from large image-text datasets~\cite{schuhmann2022laion, ordonez2011im2text, liu2024improved, sharma2018conceptual, das2017visual}, enable them to match or even surpass specialized video architectures in tasks like temporal grounding~\cite{gao2017tall, ren2024timechat, huang2024vtimellm} and video captioning~\cite{chen2023vast, liu2024valor}.

However, the sparse sampling strategy faces challenges when dealing with long videos. The limited context length of LLMs restricts the number of sampled frames, causing significant information loss and impairing the ability to capture fine-grained details in the video. 
To address this, Video-LLMs have introduced memory~\cite{song2024moviechat, he2024ma} or compression modules~\cite{ren2024timechat, weng2024longvlm}. However, most of these approaches rely heavily on large-scale pre-training or fine-tuning, failing to effectively utilize the existing capabilities of VLMs.

Our key insight is that current VLMs already possess frame-level understanding capabilities required for video tasks, but the sparse sampling strategy fails to provide sufficient perceptual information and temporal modeling capabilities.
Short video segments, containing enough motion cues, establish a temporal baseline that enables the model to acquire temporal awareness. This temporal knowledge can then be transferred to long video understanding, providing a robust solution without the need for large-scale pre-training.

Building on this insight, we propose \textbf{L}ightweight \textbf{V}ideo \textbf{C}ompression (LVC), which introduces the novel parameter-free Query-Attention Video Compression mechanism. This mechanism transforms discrete sampled features into continuous dynamic representations. Guided by the input query, LVC extracts key information from long videos, providing a more complete video representation while suppressing irrelevant details. 
By training only the alignment layer of VLMs~(approximately 100M parameters), LVC successfully evolves VLMs into video understanding models, fully leveraging the capabilities of VLMs. 
The LVC-enhanced VLMs achieve consistent relative improvements on the long video understanding benchmarks, MLVU and Video-MME. 
Compared to Video-LLMs, which rely on millions of image-text pairs and hundreds of thousands of video-text pairs for pre-training~\cite{huang2024vtimellm, shu2024video} and even more~\cite{chen2024timemarker}, LVC only requires 10k video pairs. 
Moreover, InternVL2-8B-LVC can be trained on a single H100-80G GPU in 5 hours, offering a low-cost and efficient path to high-performance video understanding models.
In conclusion, our main contributions are as follows: 
\begin{itemize}
    \item 
    \textbf{Parameter-free Query-Attention Video Compression.}  
    We propose a query-guided video dynamic representation method that enhances the temporal awareness of VLMs. It is a plug-and-play approach compatible with any VLMs that process videos as sequences of image frames.
    \item \textbf{A lightweight framework for long video understanding.} 
    We propose a lightweight framework for converting VLMs into video models. By training only the alignment layer with only 10k short videos, the frame-level image-text understanding capabilities of VLMs are fully leveraged for long video understanding tasks.
    \item \textbf{LVC-enhanced models.} 
    We introduce InternVL2-8B/26B/40B-LVC and Phi-3.5-Vision-LVC~(4B), consistently enhancing VLMs' performance on zero-shot long video understanding benchmarks MLVU and Video-MME. Specifically, InternVL2-40B-LVC achieve scores of 68.2 and 65.9 on these benchmarks, with relative improvements of 14.6\% and 7.7\%. Notably, InternVL2-40B-LVC surpass GPT-4o on MLVU.
\end{itemize}

\label{sec:intro}

\section{Related Work}
\subsection{Vision-Language Models}
VLMs aim to perform languaged-based tasks using information from the visual modality. 
Early VLMs \cite{radford2021learning} adopt contrastive learning to align vision and language embeddings. 
With the rapid development of LLMs, VLMs have evolved to utilize LLMs' capabilities to tackle complex vision-language tasks. 
These models mainly consist of three main components: a vision encoder, a modality alignment module, and an LLM backbone such as \cite{touvron2023llama, vicuna2023}.
Modality alignment strategies range from simple linear projection~\citep{liu2023visual, zhu2023minigpt, wang2024qwen2} to more complex approaches such as a Q-Former module with learnable queries~\citep{li2023blip} or cross-attention mechanisms~\citep{alayrac2022flamingo, wang2025cogvlm}. These models typically undergo large-scale pre-training and instruction tuning on extensive image-text corpora~\citep{liu2023visual, li2024omnicorpus}, yielding strong performance across a variety of vision-language benchmarks. Furthermore, \cite{chen2024internvl, wang2024qwen2} unifies single-image, multi-image, and video data, achieving significant improvements in multimodal tasks.

\subsection{Video Large Language Models}
Since LLMs rapidly become a research hotspot, many works connect LLMs with visual encoders, leveraging LLMs' language understanding and generation capabilities for video tasks. 
Early works, such as \citep{li2024mvbench}, use a video transformer as the encoder, followed by a Q-Former to compress video features. 
Later, Video-LLMs tend to utilize a pre-trained visual encoder to process a fixed number of sampled frames. \cite{luo2023valley, maaz2023video} employ a frozen vision transformer, and \cite{lin2023video} leverages \citep{zhu2023languagebind} for pre-aligning visual features.
Subsequently, Video-LLMs incorporate compression modules to process more input frames within the LLM's context length. 
\cite{ataallah2024minigpt4, xu2024pllava} use pooling to compress visual tokens, \citep{ren2024timechat} adopts a two-layer Q-Former, and \citep{jin2024chat} applies spatiotemporal clustering. 
As compression module designs evolve, \citep{shu2024video} introduces visual summarization tokens, while \citep{liu2024nvila} implements a scale-then-compress strategy. 
These approaches rely on large-scale video datasets and multi-stage training of compression modules, alignment layers, and LLMs, trading training and data costs for performance.

\subsection{Long Video Understanding}
Long video understanding faces challenges as high information density conflicts with LLMs' limited context windows. 
One approach is hierarchical processing.
\cite{lin2023mm} uses scene detection for video clipping and then employs GPT-4V to understand clip-level video.
\cite{wang2024videoagent, wu2023visual} exemplify agent-based frameworks where LLMs strategically orchestrate vision models through dynamic invocation and output synthesis.
\cite{sanders2024tv, wang2024videotree} facilitate LLM reasoning on long videos through their adaptive tree-based video representation.
Another way is focusing on long-video representation. 
\cite{song2024moviechat, he2024ma} employ the memory mechanism to iteratively store and retrieve video frame features.
\cite{li2024llama} uses context attention to represent each frame with two distinct tokens, and \cite{weng2024longvlm} applies a token merging module.
\cite{liu2024kangaroo} progressively increases resolution and input frames for long-video adaptation.
Hierarchical methods are often complex in design and fail to fully utilize video representation information, while data-driven Video-LLMs rely on high-quality video-text pairs for training, which are difficult to acquire at scale.

\label{sec:related}
\section{Methods}
\begin{figure*}[h]
  \centering
   \includegraphics[width=\linewidth]{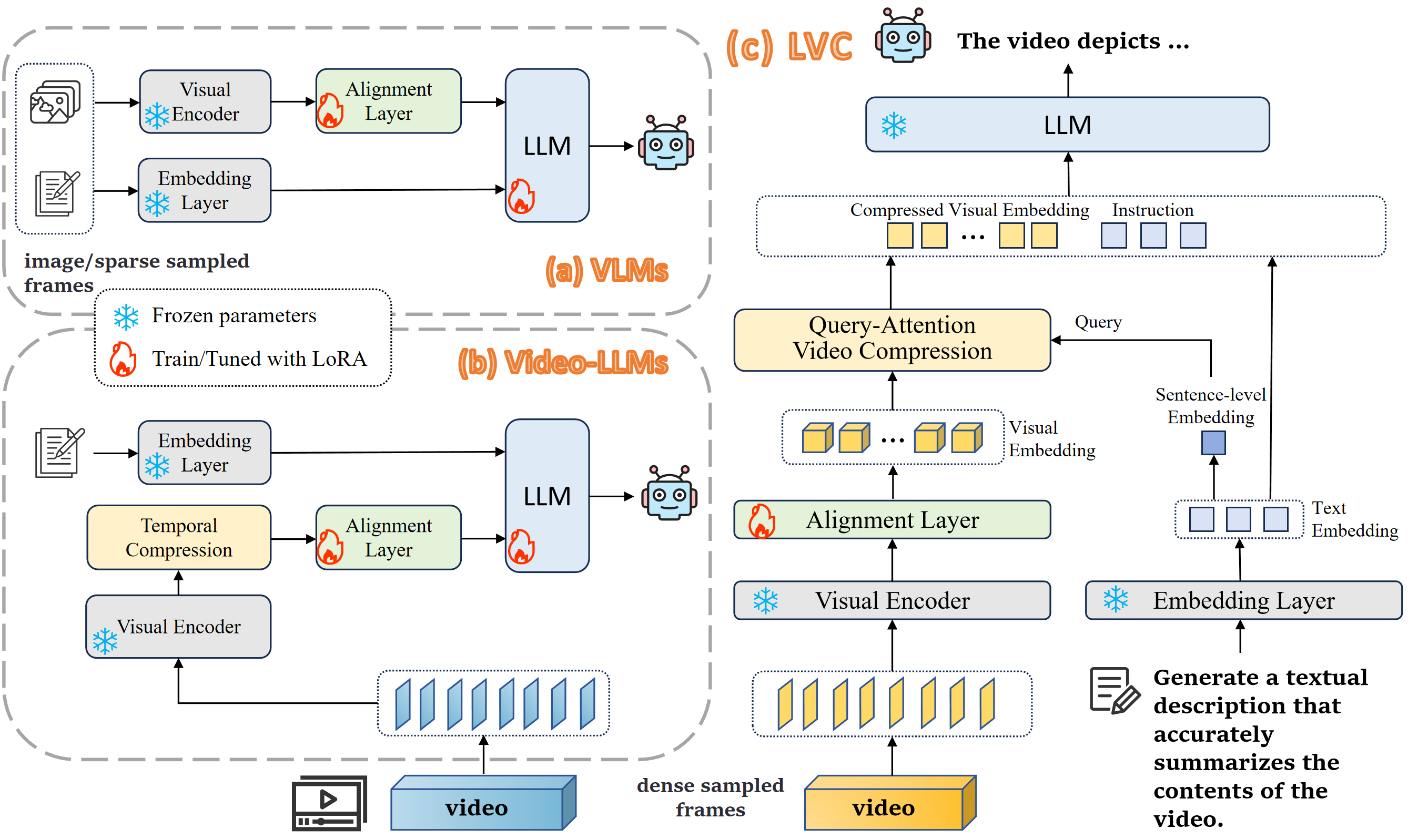} 

   \caption{(a)VLMs (b)Video-LLMs (c)The overall architecture of LVC.}
   \label{Figure2}
\end{figure*}
\subsection{Problem Formulation of Video Understanding}

General video understanding tasks can be formulated as: 
given a video $v$ and a textual query $q$, the goal is to predict a textual answer $y$.
VLMs process the video into $N$ sampled frames $f_1, f_2,...,f_N$.  
The length of visual feature tokens must account for the window size limitation of the LLM backbone. 
Therefore, in typical VLMs, $N$ is generally set to smaller values, such as 8 or 16~\cite{chen2024internvl, liu2023visual}.
These sampled frames are processed through a Vision Transformer, such as CLIP-ViT~\cite{radford2021learning}, to obtain sampled frame features $\boldsymbol{F}_1, \boldsymbol{F}_2,..., \boldsymbol{F}_N, \boldsymbol{F}_i \in \mathbb{R}^{t \times d}$. Here, $t$ represents the number of tokens describing an image frame, and $d$ denotes the feature dimension of each token.

The sampled frame features $\boldsymbol{F}_{i}, i=1,..., N$ are sequentially concatenated along the token dimension and aligned to the LLM's text space using an alignment layer. 
They are then concatenated with a query embedded as $\boldsymbol{Q} \in \mathbb{R}^{l \times d'} $ and fed into the LLM for autoregressive decoding to generate the answer.

\begin{align}
    &\boldsymbol{V} = \text{Align}([\boldsymbol{F}_1; \boldsymbol{F}_2; \dots; \boldsymbol{F}_N]) \in \mathbb{R}^{(Nt) \times d'}, \\
    &P(\boldsymbol{Y} | \boldsymbol{V}, \boldsymbol{Q}) = \prod_{j=1}^{L} P_{LLM}(\boldsymbol{Y}_j | \boldsymbol{V}, \boldsymbol{Q}, \boldsymbol{Y}_{i<j}).
\end{align}

Here, $d'$ represents the feature dimension in the LLM's text space, $\boldsymbol{V}$ represents the input visual features, and $\boldsymbol{Y}_{i<j}$ represents the text embeddings generated from previous outputs.

Video-LLMs densely sample $M$ frames and apply feature transformation to ensure that the visual features conform to the input constraints of the LLM backbone. 
The number of frames $M$ can be set to values such as 64, 128, or 256~\cite{yao2024minicpm, liu2024oryx, li2024aria}, allowing Video-LLMs to capture much more temporal information compared to VLMs. 
However, since the $M$ sampled frames undergo feature transformation to align with the input requirements of the LLM backbone, this necessitates joint training of the compression module, alignment layers, and the LLM itself. 

Instead of training the whole model with long video-text-paired data, it uses a parameter-free compression mechanism that transforms the features to closely resemble the structure of real frames—both in shape and token concatenation (including special characters). 
The compressed features are referred to as pseudo-image frames, which do not rely on extensive data-driven pre-training or fine-tuning.

\subsection{Model Architecture}

LVC is designed to extend VLMs for long video understanding with low training cost. 
As shown in Figure \ref{Figure2}(c), the overall architecture integrates multi-frame temporal information through a Visual Encoder, a parameter-free compression module, an alignment layer, and the LLM backbone. 
LVC introduces denser frame sampling to capture fine-grained temporal information and then compresses them into "pseudo-images" through the parameter-free compression module. 

Unlike the pre-training paradigms of VLMs and Video-LLMs shown in Figure \ref{Figure2}(a)(b). The only trainable component is the alignment layer, which projects compressed features into the text space of the LLM backbone. 
This design allows the well-pre-trained VLM to rapidly adapt to long video inputs without fine-tuning the LLM backbone, achieving both parameter efficiency and temporal awareness capability.

In terms of computational complexity, our method increases only the Visual Encoder’s computation, while keeping LLM-side processing unchanged, offering a novel solution for long-video understanding.

\subsection{Infusing Queries into Video Compressing}
Densely sampled frames often contain redundant information, and directly concatenating real frame features in sequence would far exceed the LLM input limitation. 
To address this, we propose the Query-Attention Video Compression mechanism, which leverages the text modality to retain essential information needed for video understanding.

Specifically, the densely sampled frames are processed through a Vision Transformer and alignment layer to obtain the visual features $\boldsymbol{V}$. 
Based on the compression ratio, a window length $w$ is set, and $\boldsymbol{V}$ is sliced as:

\begin{equation}
    \boldsymbol{V} = [\boldsymbol{V}_1,\boldsymbol{V}_2,...,\boldsymbol{V}_{M't}], \boldsymbol{V}_i \in \mathbb{R}^{w \times d'}, i=1,...,M't.
\end{equation}

$\boldsymbol{V}_i$ represents the local video features, where the video is divided into windows to ensure that the temporal information of the video is preserved.
$M' = M/w$ represents the number of pseudo-image frames after compression.

The textual query $q$ with length $l$ is passed through a Text Embedding layer to obtain the query features $\boldsymbol{Q} \in \mathbb{R}^{l \times d'}$, and then the token dimension is averaged to derive the sentence-level query feature:

\begin{equation}
    \overline{\boldsymbol{Q}} = \frac{1}{l} \sum_{j=1}^l \boldsymbol{Q}_j, \boldsymbol{Q}_j \in \mathbb{R}^{d'}.
\end{equation}

The query feature $\overline{\boldsymbol{Q}}$ is integrated into $\boldsymbol{V}$ by generating weights, which are then used to modulate the visual features:

\begin{align}
    \boldsymbol{W}_i &= \text{Softmax}\left(\frac{\overline{\boldsymbol{Q}} \boldsymbol{V}_i^\top}{\sqrt{d'}}\right) \in \mathbb{R}^{w}, \\
    \boldsymbol{V}'_i &= \sum_{k=0}^w \left( \boldsymbol{W}_i \odot \boldsymbol{V}_i \right)^{(k)} \in \mathbb{R}^{d'}, \\
    \boldsymbol{V}' &= [\boldsymbol{V}'_1, \boldsymbol{V}'_2, \ldots, \boldsymbol{V}'_{M't}] \in \mathbb{R}^{M't \times d'}.
\end{align}

The index $k$ iterates over the window size $w$, applying the corresponding weight from $\boldsymbol{W}_i$ to each token feature in $\boldsymbol{V}_i$. This process allows the model to adaptively adjust the importance of each token in the window, modulating the frame features based on their relevance to the query, thus improving the model’s ability to capture and process temporal dependencies.

Additionally, we observe that the multi-head mechanism introduces more multi-modal weight information in the compression process. 
The pseudo-code for our complete compression mechanism is provided in the supplementary material.

\label{sec:methods}
\section{Experiments}

\subsection{Dataset and Benchmarks}

\textbf{Video-ChatGPT}~\cite{maaz2023video} is a commonly used instruction-tuning dataset for general video understanding tasks, containing 13,303 videos and 100,010 video-text pairs, annotated using a combination of human-assisted and semi-automatic methods. 
The average duration of the videos is 117s (short videos). 
In our experiments, we randomly selected 10,000 pairs as a subset \textbf{Video-ChatGPT-10K} to train the alignment layer of VLMs. 
Ablation studies on the data show that LVC is not a data-driven method, which significantly distinguishes it from the pre-training paradigm.
 
\textbf{MLVU} ~\cite{zhou2024mlvu} is a multi-task long video understanding benchmark. We evaluate LVC on the multiple-choice questions ($MLVU_M$). The average length of videos is about 15 minutes, ranging from 3 minutes to 2 hours.
LVC achieved comprehensive improvements across seven sub-tasks, including action order, action count, topic reasoning, anomaly recognition, plot QA, ego reasoning, and needle QA.

\textbf{Video-MME} ~\cite{fu2024video} contains 900 videos and 2700 high-quality annotations of multiple-choice questions, including short (0-2 minutes), medium (4-15 minutes), and long videos (30-60
minutes). We evaluate the VLMs before and after enhancement with LVC without subtitles, as the sampled video frames contain more subtitles compared to the input frames of the VLMs, ensuring a fair comparison.

\subsection{Experimental Setup}

\begin{table}[ht]
    \caption{Training parameters and total parameters of InternVL2-8/26/40B-LVC and Phi-3.5-Vision-LVC, along with corresponding training time and resources.}
    \label{Table Parameters}
    \centering
    \resizebox{0.5\textwidth}{!}{
        \begin{tabular}{lcccc}
        \toprule
        Model & Train~(B) & Total~(B) & Time~(h) & Resources \\
        \midrule
        InternVL2-8B & 0.03 & 8.08 & 4.51 & 1*H100-80G \\
        InternVL2-26B & 0.12 & 25.51 & 6.31 & 2*H100-80G \\
        InternVL2-40B & 0.14 & 40.07 & 4.01 & 4*H100-80G \\
        Phi-3.5-Vision & 0.22 & 4.15 & 3.28 & 8*A100-40G \\
        \bottomrule
    \end{tabular}
    }
\end{table}

We conduct all our experiments on the Video-ChatGPT-10K dataset, which is \textbf{zero-shot} with respect to the two benchmarks. 
Considering that the average duration of the Video-ChatGPT-10K subset is less than two minutes, we fix the number of sampled frames per video to \textbf{64} (approximately 0.5 fps). 
After compression, the number of pseudo-image frames is set to \textbf{2, 4, 8}, and \textbf{16}, allowing us to comprehensively evaluate the performance of different models under varying input frame numbers.

We conduct all our experiments by training only the alignment layer of the model, with the specific training parameters and total parameter count detailed in Table \ref{Table Parameters}.
From both the data and parameter perspectives, LVC is a lightweight approach. 
Additionally, since it does not fine-tune the LLM, LVC remains a plug-and-play method, ensuring compatibility with various VLM architectures.

\subsection{Results on InternVL2 Series}

\begin{table*}[ht]
    \caption{Performance of InternVL2-8/26/40B-LVC on MLVU. TR, AR, NQA, ER, PQA, AC, AO represent the sub-tasks: topic reasoning, anomaly recognition, needle QA, ego reasoning, plot QA, action count, and action order. \textit{fps} represents the number of frames sampled per second. `$\rightarrow$16' indicates that the model maintains the same visual feature length before and after LVC enhancement for a fairer comparison.}
    \label{Table2}
    \centering
    \resizebox{0.9\textwidth}{!}{
        \begin{tabular}{lcccccccccc}
        \toprule
        Model & LLM-Size & Frames & TR & AR & NQA & ER & PQA & AC & AO & M-AVG \\
        \midrule
        Qwen-VL-MAX & - & 16 & 67.4 & 63.5 & 40.3 & 40.9 & 43.3 & 25.0 & 14.8 & 42.2 \\
        GPT-4 Turbo & - & 16 & 79.5 & 68.0 & 45.9 & 47.4 & 60.6 & 26.5 & 16.1 & 49.2 \\
        GPT-4o & - & 0.5fps & 87.4 & 74.5 & 64.8 & 57.1 & 65.1 & 56.7 & 46.3 & 64.6 \\
        \midrule
        TimeChat~\cite{ren2024timechat} & 7B & 96 & 23.1 & 27.0 & 24.5 & 28.4 & 25.8 & 24.7 & 32.0 & 30.9 \\
        Video-ChatGPT~\cite{maaz2023video} & 7B & 100 & 26.9 & 24.0 & 40.3 & 42.0 & 29.9 & 25.1 & 31.1 & 31.3 \\
        VideoChat2~\cite{li2024mvbench} & 7B & 16 & 74.6 & 51.5 & 42.0 & 47.4 & 43.8 & 22.8 & 29.6 & 44.5 \\
        MiniGPT4-Video~\cite{ataallah2024minigpt4} & 7B & 90 & 70.9 & 52.5 & 49.0 & 48.6 & 44.5 & 23.2 & 23.0 & 44.5 \\
        Video-LLaVA~\cite{lin2023video} & 7B & 8 & 71.6 & 57.0 & 53.2 & 45.2 & 48.4 & 20.1 & 35.9 & 47.3 \\
        InternVL2-76B & 70B & 16 & 87.1 & 77.0 & 65.6 & 52.0 & 65.3 & 26.2 & 46.3 & 59.9 \\
        Video-XL~\cite{shu2024video} & 7B & 256 & - & - & - & - & - & - & - & 64.9 \\
        Oryx-1.5~\cite{liu2024oryx} & 32B & 128 & - & - & - & - & - & - & - & 72.3 \\
        \midrule
        InternVL2-8B & 7B & 16 & 82.1 & 66.0 & 60.0 & 50.9 & 59.4 & 33.5 & 43.2 & 56.4 \\
        InternVL2-8B-LVC & 7B & 64$\rightarrow$16 & \textbf{84.0} & 65.0 & \textbf{68.7} & \textbf{56.0} & \textbf{69.6} & \textbf{38.8} & \textbf{50.6} & \textbf{61.8(+5.4)} \\
        \midrule
        InternVL2-26B & 20B & 16 & 87.1 & 64.0 & 59.7 & 52.3 & 63.8 & 32.5 & 41.3 & 57.3 \\
        InternVL2-26B-LVC & 20B & 64$\rightarrow$16 & 86.7 & 63.0 & \textbf{71.3} & \textbf{61.1} & \textbf{69.2} & \textbf{33.5} & 40.9 & \textbf{60.8(+3.5)} \\
        \midrule
        InternVL2-40B & 34B & 16 & 89.7 & 74.5 & 61.7 & 48.6 & 64.8 & 30.1 & 47.1 & 59.5 \\
        InternVL2-40B-LVC & 34B & 64$\rightarrow$16 & \textbf{89.0} & \textbf{76.5} & \textbf{73.5} & \textbf{60.5} & \textbf{75.7} & \textbf{43.2} & \textbf{59.1} & \textbf{68.2(+8.7)} \\
        \bottomrule
    \end{tabular}
    }
\end{table*}

\begin{table}[ht]
    \caption{Performance of InternVL2-8/26/40B-LVC on Video-MME.}
    \label{Table3}
    \resizebox{0.5\textwidth}{!}{
        \begin{tabular}{lcccccc}
        \toprule
        Model & LLM-Size & Frames & Short & Medium & Long & Overall  \\
        \midrule
        Qwen-VL-Max & - & 4 & 55.8 & 49.2 & 48.9 & 51.3 \\
        GPT-4v & - & 10 & 70.5 & 55.8 & 53.5 & 59.9 \\
        GPT-4o & - & 384 & 80.0 & 70.3 & 65.3 & 71.9 \\
        Gemini-1.5-Pro & - & 1/0.5fps & 81.7 & 74.3 & 67.4 & 75.0 \\
        \midrule
        Video-LLaVA~\cite{lin2023video} & 7B & 8 & 45.3 & 38.0 & 36.2 & 39.9 \\
        Chat-UniVi-v1.5~\cite{jin2024chat} & 7B & 64 & 45.7 & 40.3 & 35.8 & 40.6 \\
        LongVA~\cite{zhang2024long} & 7B & 128 & 61.1 & 50.4 & 46.2 & 52.6 \\
        Video-XL~\cite{shu2024video} & 7B & 128 & 64.0 & 53.2 & 49.2 & 55.5 \\
        InternVL2-76B & 70B & 16 & 72,2 & 58.0 & 53.3 & 61.2 \\
        LLaVA-Video & 72B & 64 & 81.4 & 68.9 & 61.5 & \textbf{70.6} \\
        \midrule
        MiniCPM-V2.6~\cite{yao2024minicpm} & 8B & 64 & 71.3 & 59.4 & 51.8 & 60.9 \\
        InternVL2-76B & 70B & 16 & 72.2 & 58.0 & 53.3 & 61.2 \\
        MiniCPM-o2.6 & 8B & 64 & 75.4 & 63.9 & 52.2 & 63.9 \\
        LLaVA-OneVision~\cite{li2024llava} & 72B & 32 & 76.7 & 62.2 & 60.0 & 66.3 \\
        Oryx-1.5~\cite{liu2024oryx} & 34B & 128 & 77.3 & 65.3 & 59.3 & 67.3 \\
        Qwen2-VL~\cite{wang2024qwen2} & 72B & 768 & 80.1 & 71.3 & 62.2 & \textbf{71.2} \\
        \midrule
        InternVL2-8B & 7B & 16 & 67.7 & 49.9 & 44.4 & 54.0 \\
        InternVL2-8B-LVC & 7B & 64$\rightarrow$16 & 67.0 & \textbf{51.8} & \textbf{47.3} & \textbf{55.4(+1.4)} \\
        \midrule
        InternVL2-26B & 20B & 16 & 65.8 & 50.3 & 45.4 & 53.9 \\
        InternVL2-26B-LVC & 20B & 64$\rightarrow$16 & \textbf{67.7} & \textbf{55.7} & \textbf{48.0} & \textbf{57.1(+3.2)} \\
        \midrule
        InternVL2-40B & 34B & 16 & 72.0 & 59.1 & 52.6 & 61.2 \\
        InternVL2-40B-LVC & 34B & 64$\rightarrow$16 & \textbf{74.8} & \textbf{66.0} & \textbf{56.0} & \textbf{65.9(+4.7)} \\
        \bottomrule
    \end{tabular}
    }
\end{table}

Tables \ref{Table2} and \ref{Table3} provide a detailed report on the experimental results of enhancing InternVL2~\cite{chen2024internvl} across three model scales using the LVC method. From these results, we can derive the following conclusions:
\begin{enumerate}
    \item The LVC method enhances the long-video understanding capability of the InternVL2 model series at a minimal cost by incorporating additional visual information. 
    On the MLVU benchmark, the three model scales achieve relative improvements of \textbf{9.6\%}, \textbf{6.1\%}, and \textbf{14.6\%}, respectively. 
    On the Video-MME benchmark, the relative improvements are \textbf{2.6\%}, \textbf{5.9\%}, and \textbf{7.7\%}, respectively. Notably, the InternVL2-8B-LVC outperforms the InternVL2-40B/76B models, while the InternVL2-40B-LVC  surpasses \textbf{GPT-4o} on the MLVU benchmark.
    \item The InternVL2-40B-LVC achieved consistent improvements across both benchmarks and their respective sub-tasks. 
    This can be attributed to its larger scale and more high-quality pre-training data.
    LVC effectively addresses the information sparsity of VLMs in temporal reasoning by incorporating denser visual inputs.
    \item The superior performance of LVC on MLVU compared to Video-MME can likely be attributed to the limited length of training data. 
    The number of sampled frames during training is insufficient to represent the hour-long videos in Video-MME. 
    In contrast, MLVU videos exhibit a smaller variance in duration, predominantly ranging from 5 to 15 minutes, making them more effectively represented by 64 sampled frames. 
\end{enumerate}

To the best of our knowledge, LVC is the first approach to enhance VLMs for long video understanding, introducing a novel research direction distinct from Video-LLMs. 
Its core contribution lies in leveraging the high-quality pre-training of VLMs to improve temporal reasoning without requiring extensive video-specific pre-training.

LVC preserves the pseudo-image frame format, including visual feature dimensions and special tokens. 
Since the windowing strategy along the token dimension allows similar tokens to be aggregated through the query-attention mechanism, the compressed video features remain closely aligned with real image frames. 
As a result, VLMs can learn the visual representation of pseudo-image frames solely through the alignment layer.

\subsection{Analyses on Input Frames}

\begin{figure}[h]
  \centering
   \includegraphics[width=\linewidth]{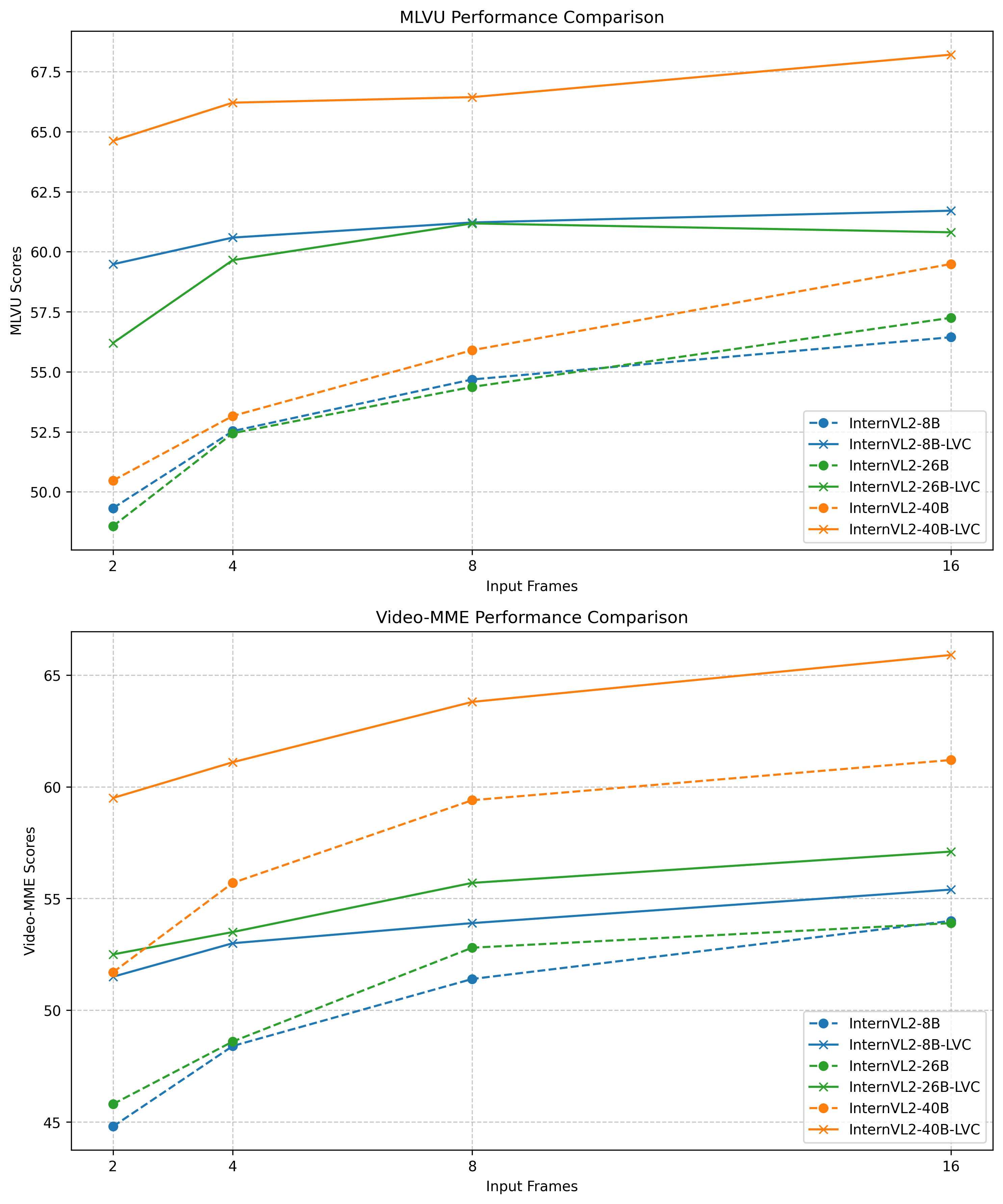}

   \caption{Performance comparison of InternVL2-8/26/40B before and after LVC enhancement on MLVU and Video-MME, with input frames set to 2/4/8/16. For LVC models, input frames refer to the numbers of pseudo-image frame after compression. Sampled frames are uniformly set to 64.}
   \label{Figure3}
\end{figure}

Figure \ref{Figure3} demonstrates the performance improvements of the LVC method in enhancing InternVL2-8B/26B/40B at compression ratios of 32/16/8/4. 
Based on these results, we can draw the following conclusions:
\begin{enumerate}
    \item The LVC method provides consistent performance improvements across various compression ratios for different scales of InternVL2, with more significant improvements observed on MLVU. Specifically, InternVL2-40B achieved relative performance improvements of 27.9\% and 24.7\% at frame=2/4, respectively.
    \item The performance improvement of LVC becomes more significant as the compression ratio increases. This is because with fewer input frames, the amount of information available to the VLMs decreases. On the other hand, it also highlights LVC's ability to retain more information even under high compression ratios.
    \item The performance growth of models enhanced with LVC becomes more gradual as the input frame count increases. For example, at frame=4/8/16, the performance differences across InternVL2 models are relatively small. This demonstrates the potential of LVC, as it is capable of providing rich and effective information even with a smaller number of input frames.
\end{enumerate}

In summary, LVC is a highly promising method for enhancing VLMs under limited computational resources. By efficiently representing pseudo-image frames, it provides the visual information required for long video understanding in VLMs.

\subsection{Ablation Studies}

\begin{table}[ht]
    \caption{Data ablation experiment on InternVL2-8/26/40B with input frames set to 16. \textbf{Origin} represents the model's original performance, while \textbf{Trained} and \textbf{LVC-trained} indicate the results of training the model with the same data, without and with LVC, respectively.}
    \label{Table4}
    \resizebox{0.5\textwidth}{!}{
        \begin{tabular}{ccccc}
        \toprule
        Model & Benchmark & Origin & Trained & LVC-trained \\
        \midrule
        InternVL2-8B & MLVU & 56.4 & 56.5 & \textbf{61.8} \\
        InternVL2-8B & Video-MME & 54.0 & 53.3 & \textbf{55.4} \\
        \midrule
        InternVL2-26B & MLVU & 57.3 & 56.7 & \textbf{60.8} \\
        InternVL2-26B & Video-MME & 53.9 & 54.8 & \textbf{57.1} \\
        \midrule
        InternVL2-40B & MLVU & 59.5 & 61.1 & \textbf{68.2} \\
        InternVL2-40B & Video-MME & 61.2 & 61.9 & \textbf{65.9} \\
        \bottomrule
    \end{tabular}
    }
\end{table}

\begin{table}[ht]
    \caption{Component ablation results of InternVL2-8B. \textbf{w/o multi-heads} indicates the removal of the multi-head mechanism in LVC, while \textbf{w/o query-attn} removes the query influence, making it equivalent to \textbf{average pooling}.}
    \label{Table5}
    \resizebox{0.5\textwidth}{!}{
        \begin{tabular}{cccccc}
        \toprule
        Model & Benchmark & Frame=2 & Frame=4 & Frame=8 & Frame=16 \\
        \midrule
        InternVL2-8B-LVC & MLVU & 59.5 & 60.6 & 61.2 & 61.8 \\
        w/o multi-heads & MLVU & 55.7 & 58.2 & 58.3 & 60.8 \\
        w/o query-attn & MLVU & 58.0 & 59.8 & 60.1 & 60.6 \\
        \midrule
        InternVL2-8B-LVC & Video-MME & 51.5 & 53.0 & 53.9 & 55.4 \\
        w/o multi-heads & Video-MME & 47.9 & 50.9 & 52.6 & 54.3 \\
        w/o query-attn & Video-MME & 47.0 & 49.6 & 50.9 & 52.9 \\
        \bottomrule
    \end{tabular}
    }
\end{table}

\vspace{1pt}
\noindent\textbf{Data Ablation:} Table \ref{Table4} presents the results of the data ablation study, where we compare the performance of the InternVL2 series models trained on Video-ChatGPT-10K with and without LVC. 
Based on the comparison between Origin and Trained, we can conclude that the LVC method does not achieve performance improvement through training on video data, but rather through its enhanced representation mechanism.

The primary role of Video-ChatGPT-10K is to help the enhanced model understand pseudo-image frame representations, as the model itself was not aligned with compressed visual features during its pre-training process.

\begin{figure}[h]
  \centering
   \includegraphics[width=\linewidth]{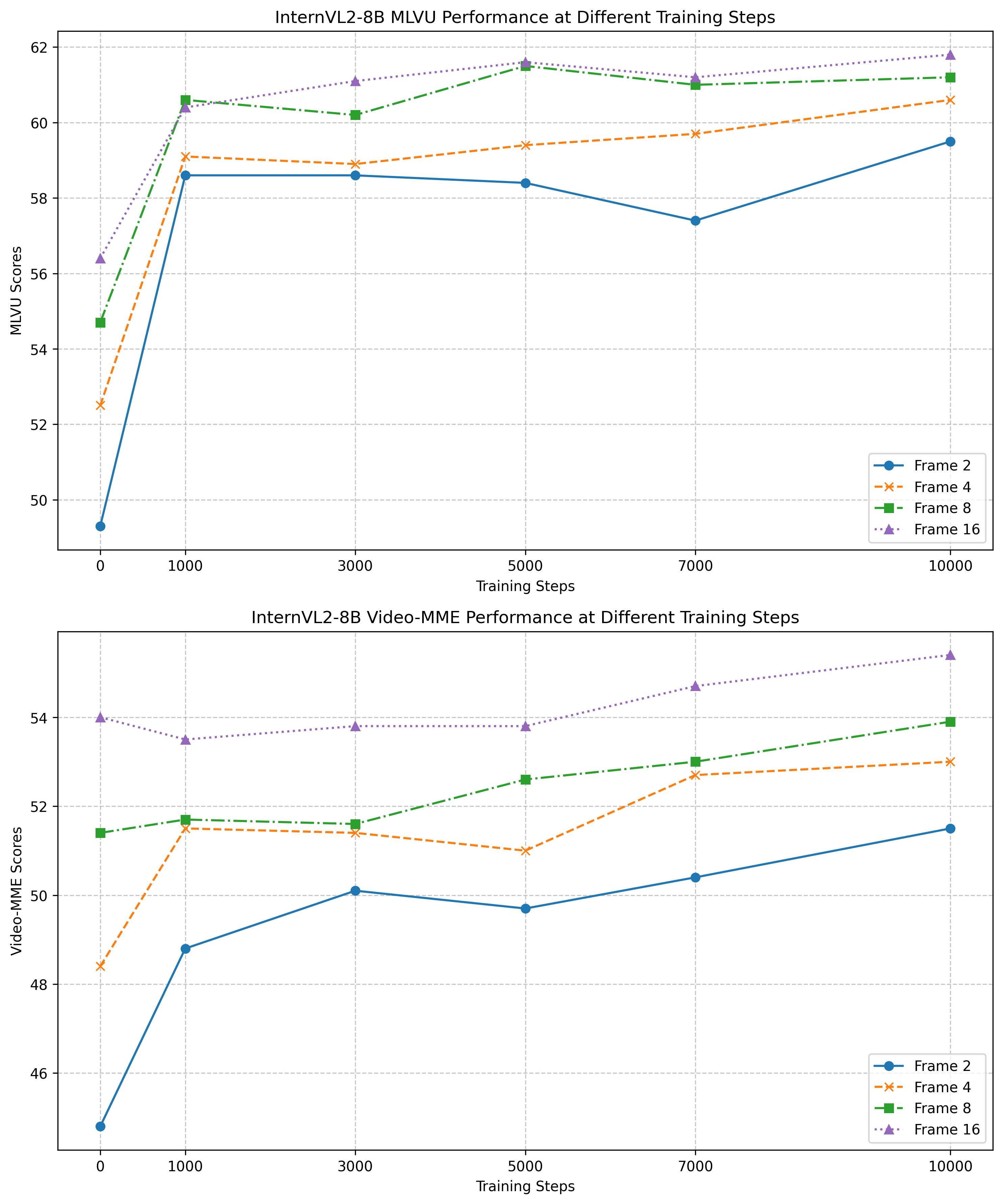}

   \caption{Performance of InternVL2-8 on MLVU with LVC enhancement at different training steps.}
   \label{Figure4}
\end{figure}

\vspace{1pt}
\noindent\textbf{Data Scale Ablation:} We add the performance of InternVL2-8B at different steps on the Video-ChatGPT10k dataset, as shown in the Figure \ref{Figure4}. We can conclude that LVC achieves quite good performance with just a small-scale dataset (around 1000 steps), indicating fast adaptation speed. 
After 1000 steps, the long video understanding ability grows slowly.

We use 10k data as a balanced choice between performance and training cost, as well as for easy comparison experiments across different models and compression ratios.

\vspace{1pt}
\noindent\textbf{Component Ablation:} Table \ref{Table5} presents the component ablation study, where we separately remove the Query-Attention Video Compression mechanism itself and the internal multi-head mechanism. The former is equivalent to average pooling. Based on the results, we can draw the following conclusions:
\begin{enumerate}
    \item The query provides multi-modal information and plays a guiding role in the video compression process of LVC.
    \item The introduction of multi-heads increases the number of weights, from one per window to multiple per window, corresponding to the number of attention heads. The finer-grained weighting has a positive correlation with performance.
    \item Even with the removal of certain components, models enhanced with LVC still achieve performance improvements over the original models in long video understanding tasks. This is because the core advantage of LVC lies in compensating for the information loss in VLMs, which rely on sparsely sampled frames as multi-image inputs.
\end{enumerate}

\subsection{Results on Phi-3.5-Vision}
\begin{figure}[h]
  \centering
   \includegraphics[width=\linewidth]{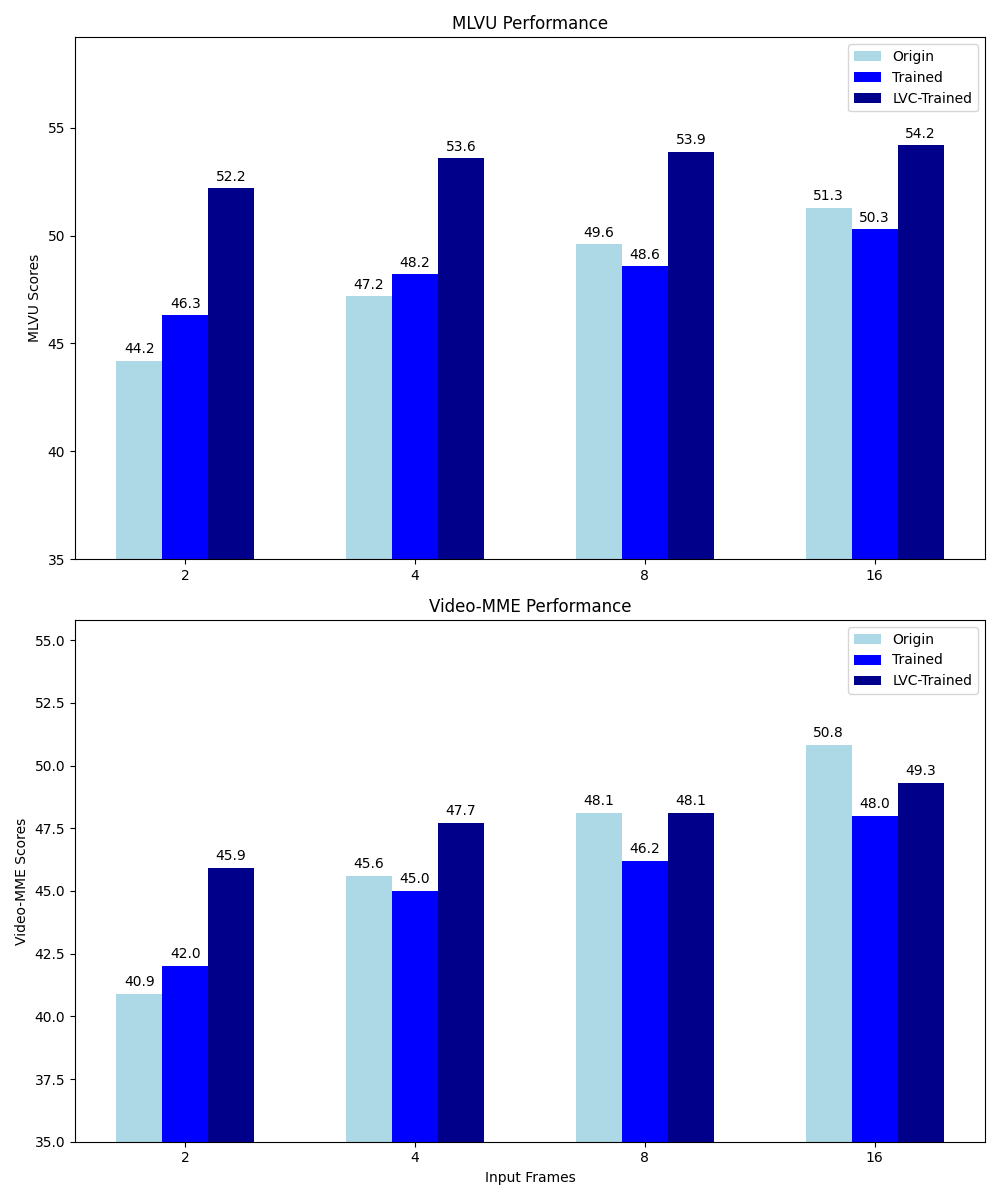}

   \caption{Performance of Phi-3.5-Vision before and after LVC enhancement, along with data ablation results~(middle).}
   \label{Figure5}
\end{figure}

We select a smaller-scale model, Phi-3.5-Vision~(4B)~\cite{abdin2024phi}, and evaluate its performance before and after enhancement with the LVC method. The results are shown in Figure \ref{Figure5}.

However, due to low dataset quality or incompatibility between the data and the model, directly training the model results in some performance degradation across both benchmarks. This phenomenon is also observed in InternVL2.

From the figure, it can be observed that LVC achieves consistent improvements on MLVU, with an average improvement of 5.4. 
Similarly, on Video-MME, LVC also demonstrates consistent improvements after excluding data influence, with an average improvement of 2.5. 
This confirms the scalability of LVC to other VLMs.

\label{sec:experiment}
\section{Discussion and Future Work}
The core idea of our proposed LVC method is to enable VLMs to process richer, continuous representations of video content. 
By introducing the Query-Attention Video Compression mechanism and representing compressed features as pseudo-image frames, we effectively reduce both data requirements and training parameters compared to conventional Video-LLMs, which rely heavily on large-scale pre-training and fine-tuning.

In contrast to Video-LLMs, our experimental results provide preliminary evidence for the effectiveness of an alternative research direction—enhancing VLMs through lightweight compression.
Moreover, by introducing additional information through denser sampling, LVC can strengthen the temporal awareness of more powerful VLMs without modifying the LLM backbone, thus offering flexibility and adaptability. 
This zero-modification aspect for LLMs allows LVC to be easily adopted in different architectures and tasks that require long video understanding.
We envision several ways to extend LVC:
\begin{enumerate}
    \item Adaptive Sampling and Compression: 
    Dynamically determine the sampling rate and compression ratio based on video length and content, optimizing the trade-off between information retention and computational cost.
    \item Advanced Compression Mechanisms: 
    Investigate more sophisticated strategies such as hierarchical compression or more robust similarity metrics to further improve compression efficiency.
    \item Multiple Alignment Layers: 
    Assign specific alignment layers to handle different video lengths or content types, potentially improving generalization across various scenarios.
\end{enumerate}

\label{sec:discussion}
\section{Conclusion}
We propose LVC and design an effective Query-Attention Video Compression mechanism to compensate for the information loss in VLMs caused by sparse sampling. Unlike common Video-LLMs, LVC is lightweight in terms of both data and training parameters. By training only the alignment layer, LVC can be easily extended to other VLMs. Extensive experiments demonstrate the superiority of LVC, offering a novel research direction for enhancing VLMs in long video understanding tasks. In the future, we will continue to extend LVC to enhance its practical applicability.

\label{sec:conclusion}
{
    \small
    \bibliographystyle{ieeenat_fullname}
    \bibliography{main}
}

\end{document}